\documentclass[11pt,a4paper]{article}

\usepackage[hyperref]{emnlp-ijcnlp-2019}
\usepackage{times}
\usepackage{latexsym}
\usepackage{graphicx, caption, subcaption}
\usepackage{url}
\usepackage{booktabs} 
\usepackage{multirow}
\usepackage[normalem]{ulem}
\useunder{\uline}{\ul}{}
\usepackage{booktabs}

\aclfinalcopy 

\title{Generation-Distillation for Efficient Natural Language Understanding in Low-Data Settings}

\author{
  Luke Melas-Kyriazi \\
  {\small {lmelaskyriazi@college.harvard.edu}} \\\And
  George Han \\
  {\small{hanz@college.harvard.edu}} \\\And
  Celine Liang \\
  {\small{cliang@college.harvard.edu}} 
  }
 
\date{}

\begin{document}
\maketitle
\begin{abstract}

Over the past year, the emergence of transfer learning with large-scale language models (LM) has led to dramatic performance improvements across a broad range of natural language understanding tasks. However, the size and memory footprint of these large LMs makes them difficult to deploy in many scenarios (e.g. on mobile phones). Recent research points to knowledge distillation as a potential solution, showing that when training data for a given task is abundant, it is possible to distill a large (teacher) LM into a small task-specific (student) network with minimal loss of performance. However, when such data is scarce, there remains a significant performance gap between large pretrained LMs and smaller task-specific models, even when training via distillation. In this paper, we bridge this gap with a novel training approach, called \textit{generation-distillation}, that leverages large finetuned LMs in two ways: (1) to generate new (unlabeled) training examples, and (2) to distill their knowledge into a small network using these examples. Across three low-resource text classification datsets, we achieve comparable performance to BERT while using $300\times$ fewer parameters, and we outperform prior approaches to distillation for text classification while using $3\times$ fewer parameters.

\end{abstract}

\section{Introduction}
Over the past year, rapid progress in unsupervised language representation learning has led to the development of increasingly powerful and generalizable language models \cite{gpt,bert}. Widely considered to be NLP's ``ImageNet moment" \citep{ruder_2018}, this progress has led to dramatic improvements in a wide range of natural language understanding (NLU) tasks, including text classification, sentiment analysis, and question answering \cite{glue,squad}. The now-common approach for employing these systems using transfer learning is to (1) pretrain a large language model (LM), (2) replace the top layer of the LM with a task-specific layer, and (3) finetune the entire model on a (usually relatively small) labeled dataset. Following this pattern, \citet{elmo}, \citet{ulmfit}, \citet{gpt}, and \citet{bert} broadly outperform standard task-specific NLU models (i.e. CNNs/LSTMs), which are initialized from scratch (or only from word embeddings) and trained on the available labeled data. 

Notably, transfer learning with LMs vastly outperforms training task-specific from scratch in low data regimes. For example, GPT-2 is capable of generating coherent text in a particular style (i.e. poetry, Java code, questions and answers) when conditioned on only a handful of sentences of that style \cite{gpt}. Similarly, on discriminative tasks such as question answering, BERT reaches accuracies comparable to previous task-specific models with orders of magnitude less labeled data \cite{bert}.

At the same time however, these large language models are extremely unwieldy. The largest versions of GPT-2 and BERT have over 1.5B and 340M parameters, respectively; it is challenging to use either of these models on a modern GPU (with 12GB of VRAM) and nearly impossible to deploy them on mobile or embedded devices.  
Thus, there is a strong need for efficient task-specific models that can leverage the knowledge from large pretrained models, while remaining highly compressed. 

\begin{figure*}[h!]
    \centering{
        \includegraphics[width=\linewidth]{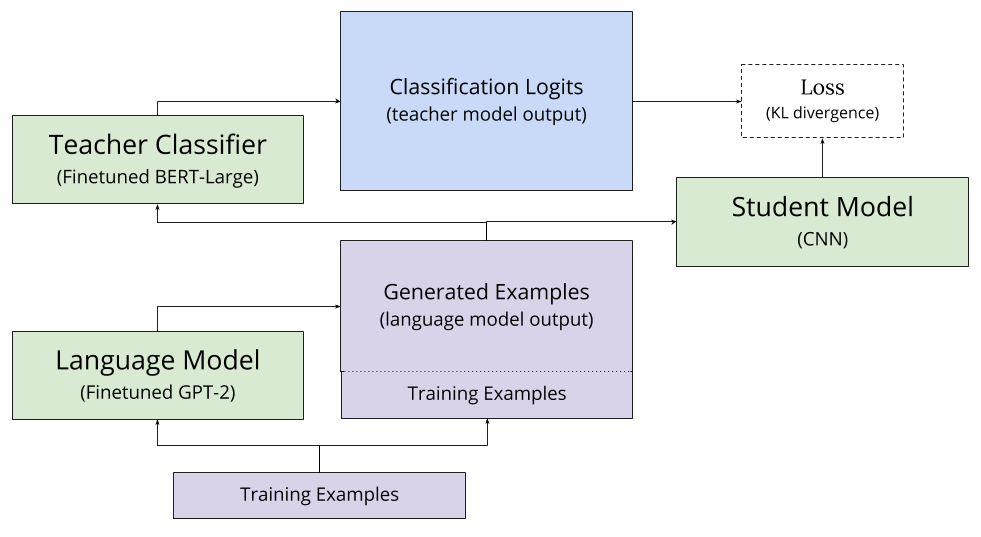}
    }
    \caption{Our proposed generation-distillation training procedure. First, we use a large language model to augment our set of training examples, and second we train our student via distillation with a large language model-based classifier. In the diagram above, green blocks indicate models and purple blocks indicate text data. }
    \label{fig:model}
\end{figure*}

In this project, we attempt to bridge this gap for the task of low-resource text classification. We propose a new approach, called \textit{generation-distillation}, to improve the training of small, task-specific text classification models by utilizing multiple large pretrained language models. First, we use a large LM (GPT-2) to generate text in the style of our training examples, augmenting our data with unlabeled synthetic examples. Second, we use the synthetic examples to distill a second large LM (BERT), which has already been finetuned for classification, into a small task-specific model (CNN).

In our experiments, we show that this procedure delivers significant gains over a standard distillation approach in low-data regimes. Specifically, on low-data versions of three widely-adopted text classification datasets (AG News, DBPedia, Yahoo Answers), we obtain 98\% of BERT's performance with 300$\times$ fewer parameters. Moreover, compared to prior work on distilling BERT \cite{t2cnn} on these datasets, we outperform past approaches while using $3\times$ fewer parameters.

\section{Related Work}

\begin{table*}[t] 
\def\arraystretch{1.25}
\begin{tabular}{l| c |c |c |c }
                        \textit{Model} & \textit{Params (1000s)}    & \textit{AG News}             & \textit{DBPedia}                                & \textit{Yahoo Answers}                          \\ \hline \hline
Baseline - TFIDF + SVM \cite{tfidf}                  & 18.1                                  & 81.9                         & 94.1                                       & 54.5                                               \\ \hline
Baseline - FastText \cite{fasttext}                  & N/A                                   & 75.2                         & 91.0                                       & 44.9                                               \\ \hline \hline 
BERT-Large                                           & 340,000                               & {\ul \textit{89.9}}          & \textbf{97.1}                              & \textbf{67.0}                                      \\ \hline \hline 
\citet{t2cnn} - BlendCNN*                            & 3617                                  & 87.6                         & 94.6                                       & 58.3                                               \\ \hline
\citet{t2cnn} - BlendCNN + \textit{Dist}*$\quad$     & 3617                                  & \textbf{89.9}                & 96.0                                       & 63.4                                               \\ \hline
Ours (Kim-style)                                     & 1124                                  & 85.7                         & 94.3                                       & 62.4                                               \\ \hline
Ours (Res-style)                                     & 1091                                  & 86.2                         & 94.7                                       & 60.9                                               \\ \hline
Ours + \textit{Dist} (Kim-style)                     & 1124                                  & 86.9                         & 95.0                                       & 62.9                                      \\ \hline
Ours + \textit{Dist} (Res-style)                     & 1091                                  & 87.3                         & 95.4                                       & 62.2                                                    \\ \hline
Ours + \textit{Gen-Dist} (Kim-style)                 & 1124                                  & {\ul \textit{89.9}}          & {\ul \textit{96.3}}                        & 64.2                                                     \\ \hline
Ours + \textit{Gen-Dist} (Res-style)                 & 1091                                  & 89.8                         & 96.0                                       & {\ul \textit{65.0}}
\end{tabular}

\caption{ \textit{(Results)} A comparison of model size and accuracy on $3$ text classification datasets. Bold font indicates best accuracy and italics+underline indicates second-best accuracy. Generation-distillation broadly improves small model performance over distillation, which in turn broadly improves performance over training from scratch. * results from other papers. }
\label{fig:results}
\end{table*}

Designed to produce contextual word embeddings, large language models (LMs) build upon the now-classic idea of using pretrained word embeddings to initialize the first layer of deep natural language processing models \cite{nlp_from_almost_scratch}. Early proponents of contextual word vectors, including CoVe, ULMFit, and ELMo \cite{cove, ulmfit, elmo}, extracted word representations from the activations of LSTMs, which were pretrained for either machine translation (CoVe) or for language modeling (ULMFit, ELMo). 

Recent work has adopted the transformer architecture for large-scale language representation. BERT \cite{bert} trains a transformer using masked language modeling and next sentence prediction objectives, giving state-of-the-art performance across NLU tasks. GPT/GPT-2 \cite{gpt} trains a unidirectional objective, showing the ability to generate impressively coherent text. 

Due to the unwieldy size of these models, a line of recent research has investigated how to best compress these models \cite{distillingbert}. In the most popular of these approaches, knowledge distillation \cite{original_distillation_paper}, the outputs of a larger ``teacher'' model are used to train a smaller ``student'' model. These outputs may contain more information than is available in the true label, helping bring the performance of the student closer to that of the teacher. On the task of text classification,  \cite{distillingbert} and \cite{t2cnn} both recently showed that it is possible to compress transformer-based LMs into CNNs/LSTMs with fewer parameters, at the cost of a small (but nontrivial) drop in accuracy. 

Our project builds on prior work in multiple ways. When performing generation-distillation, we employ a finetuned GPT-2 \cite{gpt} as our generator and a finetuned BERT \cite{bert} as our teacher classifier. Additionally, the distillation component of our generation-distillation approach is similar to the method used in \cite{t2cnn}, but with a different loss function (KL divergence in place of mean absolute error).

\section{Methodology}

As shown in Figure \ref{fig:model}, our \textit{generation-distillation} approach is divided into three steps: finetuning, generation and distillation.

\subsection{Finetuning}

The first step in our approach involves finetuning two different large LMs on our small task-specific dataset. First, we finetune a generative model (in our case, GPT-2) using only the text of the dataset. This model is used to generate new synthetic examples in the \textit{generation} step. Second, we finetune a large LM-based classifier (in our case, BERT with an added classification head) using both the text and the labels of the dataset. This model is used as the teacher in the \textit{distillation} step. 

\subsection{Generation}

In the generation step, we used a large generative LM, finetuned in the first step, to augment our training dataset with synthetic examples. Specifically, we use GPT-2 to generate new sentences in the style of our training dataset and add these to our training dataset. We do not have labels for these generated sentences, but labels are not necessary because we train with distillation; our goal in generating synthetic examples is not to improve the large LM-based classifier, but rather to improve our ability to distill a large LM-based classifier into a small task-specific classifier. 

\subsection{Distillation}
We combine both the real training examples and our synthetic examples into one large training set for distillation. We distill a large LM-based teacher classifier, finetuned in the first step, into our smaller student model via standard distillation as in \citet{original_distillation_paper}. For our loss function, like  \citet{original_distillation_paper}, we use the KL divergence between the teacher logits and the student logits; this differs from \citet{t2cnn}, who use the mean absolute error between the logits. 

\section{Experiments}

\subsection{Data}
We perform text classification on three widely-used datasets: \textit{AG News}, \textit{DBPedia}, and \textit{Yahoo Answers} \citep{agnews,dbpedia,yahoo_answers}. For purposes of comparison, we select our training set using the same procedure as \citet{t2cnn}, such that the training set contains 100 examples from each class. For the generation-distillation experiments, we use GPT-2 to generate $13600$ synthetic training examples on AG News and $25000$ synthetic training examples on DBPedia and Yahoo Answers. Combining these with the $400, 1400$, and $1000$ original (labeled) examples yields a total of $14000, 26400,$ and $26000$ examples on AG News, DBPedia, and Yahoo Answers, respectively. 

\subsection{Finetuning Details and Examples}
We finetune GPT-2 345M using Neil Shepperd's fork of GPT-2:  {\small \url{https://github.com/nshepperd/gpt-2/blob/finetuning/train.py}}

Finetuning is performed for a single epoch with a learning rate of $2e-5$ with the Adam optimizer. We use batch size 1 and gradient checkpointing in order to train on a single GPU with 12GB of VRAM. We choose to train for only 1 epoch after examining samples produced by models with different amounts of finetuning; due to the small size of the dataset relative to the number of parameters in GPT-2, finetuning for more than 1 epoch results in significant dataset memorization. 

For sampling, we perform standard sampling (i.e. sampling from the full output distribution, not top-p or top-k sampling) with temperature parameter $T=1$. Although we do not use top-k or top-p sampling, we believe it would be interesting to compare the downstream effect of different types of sampling in the future.

In Supplementary Table 3, we show examples of synthetic training texts generated by sampling from the finetuned GPT-2 model, for both DBPedia and Yahoo Answers. 

In Supplementary Table 4, we show two synthetic training texts along with their nearest neighbors in the training set. Nearest neighbors were calculated by ranking all examples from the training dataset (1400 examples) according to cosine similarity of TF-IDF vectors. As can be seen in the example in the right column, the GPT-2 language model has memorized some of the entities in the training dataset (i.e. the exact words ``Ain Dara Syria''), but provides a novel description of the entity. This novel description is factually incorrect, but it may still be helpful in training a text classification model in a low-resource setting, because the words the model generates (i.e. ``Syria'', ``Turkey'', ``Karzahayel'') are broadly related to the original topic/label. For example, they may help the model learn the concept of the class ``village'', which is the label of Nearest Neighbor 1.

\begin{figure*}[h]
\centering
\includegraphics[width=0.8\textwidth]{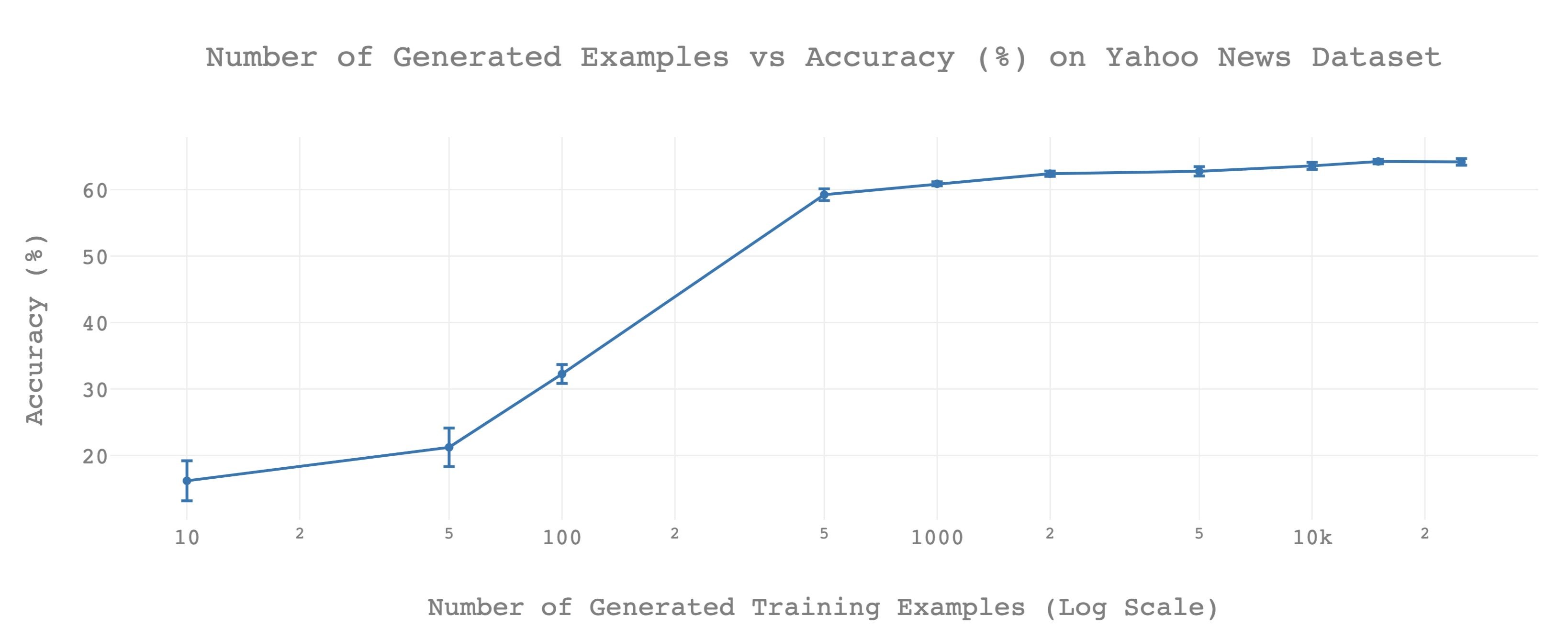}
\caption{Above, we show how the accuracy of the final distilled model varies with the number of synthetic training examples generated by GPT-2. Error bars show the standard deviation of accuracies on five separate runs. The same GPT-2 model (trained on 100 examples per class, or a total of 1000 examples) was used to generate all synthetic texts.}
\end{figure*}

\begin{table*}[t]
\centering 
\vspace{5mm}
{ \large Hard Labeling vs. Distillation on Generated Examples (Yahoo Answers)} 
\vspace{1.5mm}

\begin{tabular}{l|c|c}
         & Hard Labeling with BERT      & Distillation with BERT \\ \hline
Accuracy & 62.9 $\pm$ 0.22 & 64.2 $\pm$ 0.13 \\
\end{tabular}

\caption{Above, we show a comparison of hard labeling and distillation for labeling the synthetic examples produced by our generator network. We report the the mean and standard error of the student (Kim) model accuracy across 5 random restarts on the Yahoo Answers dataset. Generation and distillation significantly outperforms generation and hard labeling. 
}
\end{table*}

\subsection{Student Models \& Optimization}
We experiment with two main CNN architectures. The first is a standard CNN architecture from \citet{kimcnn}. The second is a new CNN based on ResNet \cite{resnet}. This ``Res-style'' model has 3 hidden layers, each with hidden size 100, and dropout probability $p=0.5$. We use multiple models to demonstrate that our performance improvements over previous approaches are not attributable to architectural changes, and to show that our approach generalizes across architectures. 

We train the CNNs using Adam \citep{adam,adamw} with learning rate $10^{-3}$. Additionally, the CNNs both use 100-dimensional pretrained subword embeddings \citep{bpemb}, which are finetuned during training.

\subsection{Results}

We report the performance of our trained models in Table \ref{fig:results}.

When trained with standard distillation, our KimCNN and ResCNN models perform as would be expected given the strong results in \citet{t2cnn}. Our models perform slightly worse than the 8-layer BlendCNN from \citet{t2cnn} on AG News and DBPedia, while performing slightly better on Yahoo Answers. Standard distillation improves their performance, but there remains a significant gap between the CNNs and the BERT-Large based classifier. Training with the proposed generation-distillation approach significantly reduces the gap between the CNNs and BERT-Large; across all datasets, the model trained with generation-distillation matches or exceeds both the model the model trained with standard distillation and the BlendCNN. 

\subsection{Ablation}

In Figure 2, we show how the accuracy of the final distilled model varies with the number of synthetic training examples generated by GPT-2. The distilled model is trained entirely on synthetic examples, without ever seeing the original data. The model shows strong performance (60\% accuracy) with as few as 500 generated training examples, or $50$ per class. Moreover, model performance continues to increase with more generated training examples, up to $25,000$. 

In Table 2, we compare two different methods of labeling the synthetic examples produced by our generator network (GPT-2): hard labeling and distillation. Hard labeling refers to taking the maximum-probability class according to our finetuned BERT model as the label for each generated example and using a standard cross entropy loss function. Distillation refers to using the probability distribution outputted by BERT as the label for each generated examtple and using a KL divergence loss function. Put differently, in the former we use BERT to generate  labels, whereas in the latter we use BERT to generate perform distillation. We find that generation and distillation outperforms generation and hard labeling by a significant margin, consistent with previous work on knowledge distillation \cite{original_distillation_paper}.

\section{Conclusion}

In this work, we present a new approach to compressing natural language understanding models in low-data regimes. Our approach leverages large finetuned language models in two ways: (1) to generate new (unlabeled) training examples, and (2) to distill their knowledge into a small network using these examples. Across three low-resource text classification datsets, we achieve comparable performance to BERT while using $300\times$ fewer parameters, and we outperform prior approaches to distillation for text classification while using $3\times$ fewer parameters. Although we focus on text classification in this paper, our proposed method may be extended to a host of other natural language understanding tasks in low-data settings, such as question answering or extractive summarization.

\bibliography{emnlp-ijcnlp-2019}
\bibliographystyle{acl_natbib}

\newpage

\def\arraystretch{1.4}
\begin{table*}[t]
\centering
{ \large Examples of Generated Training Texts } 

\vspace{1.5mm}

\begin{tabular}{|p{0.95\textwidth}|}
\hline
\hline
\multicolumn{1}{|c|}{\textbf{DBPedia}}                                                                                                                                                                                                                                                                                                                                                                           \\ \hline
Landmine: Landmine{[}1{]} (also known as LNG mine) is a landmine created by the Chernobyl nuclear powerplant. It is a slurry subterranean mine typically laid in shallow pools of water. The mines are connected by run-off points and can be faced off against one another.                                                                                                                            \\ \hline
Naukembe Consolidated School: Naukembe is a boarder boarding and lodging school based in the township of Naushere East Sussex England. The school is a member of the N30 co-education network. The school holds around 750 students from grade six to eleven.                                                                                                                                           \\ \hline
Peter Moldegayrd:  Peter Moldegayrd (born 6 July 1940) is a Swedish film director known for his 1972 Melancholia. He later worked in Zurich and Hong Kong. \\ \hline
Ain Dara Syria: Ain Dara (Arabic: Andin Qasim Qasim; also Romanized as Andin Qasi Qasim and Maididi Dariqat) is a small village in Doubs Governorate southwestern Syria close to the Turkey-Syria border. Nearby localities include Afrin to the north Karzahayel to the east and Siloamfara to the northwest. Ain Dara is settled by around 30 families. \\ \hline \hline
\multicolumn{1}{|c|}{\textbf{Yahoo Answers}}                                                                                                                                                                                                                                                                                                                                                                     \\ \hline
Why is America the most geographically illiterate first world country?                                                                                                                                                                                                                                                                                                                                  \\ \hline
Where I can get program that erases voice from music track?: Where I can get program that erases voice from music track? nowhere                                                                                                                                                                                                                                                                        \\ \hline
does anyone know the name of the song that's used in the ADIDAS commercial JosÃ© +10? (That's adidas, by the way)?: This commercial was recently in a recent adidas commercial, and they apparently used the credits for the commercial, so I saw it and thought it was pretty cool.                                                                                                                     \\ \hline
What would be a good way to express how you feel about another person?: say something nice, thoughtful, creative, professional... whatever . just let it go and move on, someone else will take care of the rest                                                                                                                                                                                        \\ \hline
\end{tabular}

\caption{ Examples of captions generated by GPT-2 for the DBPedia and Yahoo Answers datasets. The GPT-2 model that generated these texts was trained on 100 examples per class, or a total of 1000 examples for Yahoo Answers and 1400 for DBPedia. These examples were picked randomly from all generated sentences. 
}

\end{table*}

\begin{table*}[th]
\centering 
{ \large Generated Training Examples and their Nearest Neighbors in the Real Training Data (DBPedia)} 
\vspace{1.5mm}

\begin{tabular}{|p{0.1\textwidth}|p{0.4\textwidth}|p{0.4\textwidth}|}
\hline
\textbf{Generated Example}  & {\small Naukembe Consolidated School: Naukembe is a boarder boarding and lodging school based in the township of Naushere East Sussex England. The school is a member of the N30 co-education network. The school holds around 750 students from grade six to eleven. }                                                                                                                                                                                                                                        & Ain Dara {\small Syria: Ain Dara (Arabic: Qasim Qasim; \textit{\{unicode\}} also Romanized as Qasim and \textit{\{unicode\}}) is a small village in Doubs Governorate southwestern Syria close to the Turkey-Syria border. Nearby localities include Afrin to the north Karzahayel to the east and Siloamfara to the northwest. Ain Dara is settled by around 30 families. }                                                                                                                                                                 \\ \hline
\textbf{Nearest Neighbor 1} & { \small East High School (Erie Pennsylvania): East High School part of the Erie City School District is a public high school located in Erie Pennsylvania United States. The school colors are scarlet and gray. The school mascot is a Native American Warrior. People associated with East High may be referred to as East High School Warriors East High Warriors or Warriors. }                                                                                                                           & { \small Ain Dara Syria: Ain Dara (Arabic: \textit{\{unicode\}} also spelled Ayn Darah) is a small village in northern Syria administratively part of the Afrin District of the Aleppo Governorate located northwest of Aleppo. Nearby localities include Afrin to the north Karzahayel to the east and Bassouta to the south. According to the Syria Central Bureau of Statistics (CBS) Ain Dara had a population of 248 in the 2004 census.The modern-day settlement of Ain Dara is situated just east of the ancient Ain Dara temple.} \\ \hline
\textbf{Nearest Neighbor 2} & { \small Calvert School: Calvert School is a lower and middle co-educational private school with a day school operation in Baltimore Maryland and an associated homeschooling division that administers a curriculum shipped to families around the United States and the world. Developed in 1906 the home school curriculum grew by being advertised in the National Geographic magazine as a kindergarten program for those wishing to offer a better education to their children.}                         & {\small Carabus hemprichi:,Carabus hemprichi is a species of black-coloured ground beetle in the Carabinae subfamily that can be found in Israel Lebanon Syria and Turkey}                                                                                                                                                                                                                                                                                                                                                                        \\ \hline
\textbf{Nearest Neighbor 3} & { \small South Elgin High School: South Elgin High School (SEHS) opened 2004 is a four-year high school located in South Elgin Illinois a north-west suburb of Chicago Illinois in the United States. It is part of Elgin Area School District U46 which also includes Elgin High School Larkin High School Bartlett High School and Streamwood High School. The class of 2008 was the first to graduate at the high school. The class of 2009 was the first four year graduating class from the high school.} & {\small Retowy:,Retowy (German: Rettauen) is a village in the administrative district of Gmina Sepopol within Bartoszyce County Warmian-Masurian Voivodeship in northern Poland close to the border with the Kaliningrad Oblast of Russia. It lies approximately 10 kilometres (6 mi) north-west of Sepopol 14 km (9 mi) north-east of Bartoszyce and 68 km (42 mi) north-east of the regional capital Olsztyn.Before 1945 the area was part of Germany (East Prussia).}                                                            \\ \hline
\end{tabular}

\caption{Above, we show two example sentences from DPedia along with their nearest neighbors from the training dataset (DBPedia). Nearest neighbors were calculated by selecting the three examples from the training dataset (1400 examples) with the greatest TF-IDF vector cosine distance to the generated example. Note that in the above examples, a handful of unicode characters that could not be rendered in LaTeX were replaced by \textit{\{unicode\}}. 
}

\end{table*}

\end{document}